# A Cognitive Approach to Real-time Rescheduling using SOAR-RL

Juan Cruz Barsce[1], Jorge Palombarini[1,2], Ernesto Martinez[3]

[1] DEPARTAMENTO DE SISTEMAS (UTN) - Av. Universidad 450, X5900 HLR, Villa María, Argentina.
`jbarsce@frvm.utn.edu.ar`
[2] GISIQ(UTN) - Av. Universidad 450, X5900 HLR, Villa María, Argentina.
`jpalombarini@frvm.utn.edu.ar`
[3] INGAR(CONICET-UTN), Avellaneda 3657, S3002 GJC, Santa Fe, Argentina.
`ecmarti@santafe-conicet.gob.ar`

**Abstract.** Ensuring flexible and efficient manufacturing of customized products in an increasing dynamic and turbulent environment without sacrificing cost effectiveness, product quality and on-time delivery has become a key issue for most industrial enterprises. A promising approach to cope with this challenge is the integration of cognitive capabilities in systems and processes with the aim of expanding the knowledge base used to perform managerial and operational tasks. In this work, a novel approach to real-time rescheduling is proposed in order to achieve sustainable improvements in flexibility and adaptability of production systems through the integration of artificial cognitive capabilities, involving perception, reasoning/learning and planning skills. Moreover, an industrial example is discussed where the SOAR cognitive architecture capabilities are integrated in a software prototype, showing that the approach enables the rescheduling system to respond to events in an autonomic way, and to acquire experience through intensive simulation while performing repair tasks.

## 1 Introduction

In recent years, effective control of production systems is becoming increasingly difficult, because of growing requirements with regard to flexibility and productivity as well as a decreasing predictability of environmental conditions at the shop-floor. This trend has been accompanied by uncertainties in terms of an increasing number of products, product variants with specific configurations, large-scale fluctuations in demand and priority dispatching of orders. In order to face global competition, mass customization and small market niches, manufacturing systems must be primarily collaborative, flexible and responsive [1] without sacrificing cost effectiveness, product quality and on-time delivery in the presence of unforeseen events like equipment failures, quality tests demanding reprocessing operations, rush orders, delays in material inputs from previous operations and arrival of new orders [2]. In this context, reactive scheduling has becomes a key element of any real-time disruption management strategy, because the aforementioned conditions cause production schedules and plans ineffective after a short time at the shop floor, whereas at the same time opportunities arise for improving shop-floor performance based on the situation encountered after a disruption [2].

Most modern approaches to tackle the rescheduling problem involve some kind of mathematical programming to exploit peculiarities of the specific problem structure, bearing in mind prioritizing schedule efficiency [3] or stability [4]. More recently, Gersmann and Hammer [5] have developed an improvement over an iterative schedule repair strategy using Support Vector Machines. Nevertheless, feature based representation is very inefficient for generalization to unseen states, the repairing logic is not clear to the end-user, and knowledge transfer to unseen scheduling domains is not feasible [6]. In this way, many researchers have identified the need to develop interactive rescheduling methodologies in order to achieve higher degrees of flexibility, adaptability and autonomy of manufacturing systems [1, 2, 7]. These approaches require the integration of human-like cognitive capabilities along with learning/reasoning skills and general intelligence in rescheduling components. Therefore, they can reason using substantial amounts of appropriately represented knowledge, learn from its experience, explain itself and be told what to do, be aware of its own capabilities and reflect on its own behavior, and respond robustly to surprise [8]. In this way, embodying continuously real-time information from the shop-floor environment, the manufacturing system and the individual product configuration through abstract sensors, rescheduling component can gain experience in order to act and decide in an autonomic way to counteract abrupt changes and unexpected situations via abstract actuators [9, 10].

In this work, a novel real-time rescheduling approach which resorts to capabilities of a general cognitive architecture and integrates symbolic representations of schedule states with (abstract) repair operators is presented. To learn a near-optimal policy for rescheduling using simulated schedule state transitions, an interactive repair-based strategy bearing in mind different goals and scenarios is proposed. To this aim, domain-specific knowledge for reactive scheduling is developed and integrated with SOAR cognitive architecture learning mechanisms like chunking and reinforcement learning via long term memories [11]. Finally, an industrial example is discussed showing that the approach enables the scheduling system to assess its operation range in an autonomic way, and to acquire experience through intensive simulation while performing repair tasks in production schedules.

## 2 Real-time Rescheduling in SOAR Cognitive Architecture

In this approach, knowledge about heuristics for repair-based scheduling to deal with unforeseen events and disturbances are generated and represented resorting to using a schedule state simulator connected with the SOAR cognitive architecture [12]. In the simulation environment, an instance of the schedule is interactively modified by the system using a sequence of repair operators suggested by SOAR, until a repair goal is achieved or the impossibility of repairing the schedule is accepted. SOAR's solves the problem of generating and encoding rescheduling knowledge using a general theory of computation, which is based on goals, problem spaces, states and operators, which will be explained later in detail. To implement the proposed approach, the cognitive architecture is connected with the rescheduling component via .NET wrappers. In-formation about the actual schedule state comes in via the perception module which is related to an InputLink structure, and is held in the perceptual short-term memory. Symbolic first order schedule state structures in the form of id-attribute-value are extracted from InputLink, and added to SOAR's working memory. Working memory acts as a global short-term memory that cues the retrieval of rescheduling knowledge from Soar's symbolic long-term memories, as well as being the basis for initiating schedule repair actions. The three long-term symbolic memories are independent, and each one of them uses separate learning mechanisms. Procedural long-term memory is responsible for retrieving the knowledge that controls the processing; such knowledge is represented as *if-then* production rules, which match conditions against the contents of working memory, and perform their actions in parallel. Production rules can modify the working memory (and therefore the schedule state). To control rescheduling behavior, these rules generate preferences, which are used by the decision procedure to select a schedule repair operator. Operators are a key component in this approach, because they can be applied to cause persistent changes to the working memory. The latter has reserved areas that are monitored by other memories and processes, whereby changes in working memory can initiate retrievals from semantic and episodic memory, or initiate schedule repair actions through the abstract actuator in the environment. Semantic memory stores general first order structures that can be employed to solve new situations i.e. if in schedule state s1 the relations `precedes(task1,task2)` and `precedes(task2,task3)` which share the parameter object "task2" are verified, semantic memory can add the abstract relation `precedes(A,B), precedes(B,C)` to generalize such knowledge. On the other hand, episodic memory stores streams of experience in the form of *state-operator...state-operator* chains. Such knowledge can be used to predict behavior or environmental dynamics in similar situations or envision the schedule state outside the immediate perception using experience to predict outcomes of possible courses of actions when repairing a schedule. Moreover, this approach uses two specific learning mechanisms associated with SOAR's procedural memory, i.e. *chunking* and *reinforcement learning* [13], for learning new production rules as the schedule repair process is performed, and tuning the repair actions of rules that creates preferences for operator selection. Finally, repair operators suggested by the SOAR decision procedure affect the schedule state and are provided to the real-time rescheduling component via the OutputLink structure.

## 3 Schedule States, Repair Operators and Rescheduling Goals representation in SOAR

SOAR's working memory holds the current schedule state, and it is organized as a connected graph structure (a semantic net), rooted in a symbol that represents the state. The non-terminal nodes of the graph are called identifiers, the arcs are called attributes, and the values are the other nodes. The arcs that share the same node are called objects, so that an object consists of all the properties and relations of an identifier. A state is an object along with all other objects which are substructures of it, either directly or indirectly. In the upper section of Figure 1, an example of a state named `<s>` in SOAR's working memory is shown. In this figure, some details have been omitted for the sake of clarity, so there are many substructures and arcs in the state which are not shown.

In the situation depicted in Figure 1, we can see that in the working memory the actual state is called `<s>`, and it has attributes like avgTard with value 2.5 h., initTardiness with value 28.5 h. and a totalWIP of 46.83, among others. Also, there exists a resource attribute, whose value is another identifier, so it links state `<s>` with another objects called `<r1>`, `<r2>` and `<r3>`. `<r1>` corresponds to resource of the type extruder, it can process products of type A and B, with an accumulated tardiness in the resource of 17 h., and it has three tasks assigned to it: `<t1>`, `<t11>` and `<t3>`. These tasks are in turn objects in such a way that for instance `<t1>` has proper attributes to describe it like Product Type, Previous Task, Next Task, Quantity, Duration, Start, Finish and Due Date among others.

Similarly to tasks, resources have an important attribute called Processing Rate that determines the total quantity of a given type of product that the resource can process in one unit of time. Finally, a state attribute called Calculations stores data that is relevant to calculate tasks tardiness, resource tardiness, schedule total tardiness, and derived values.

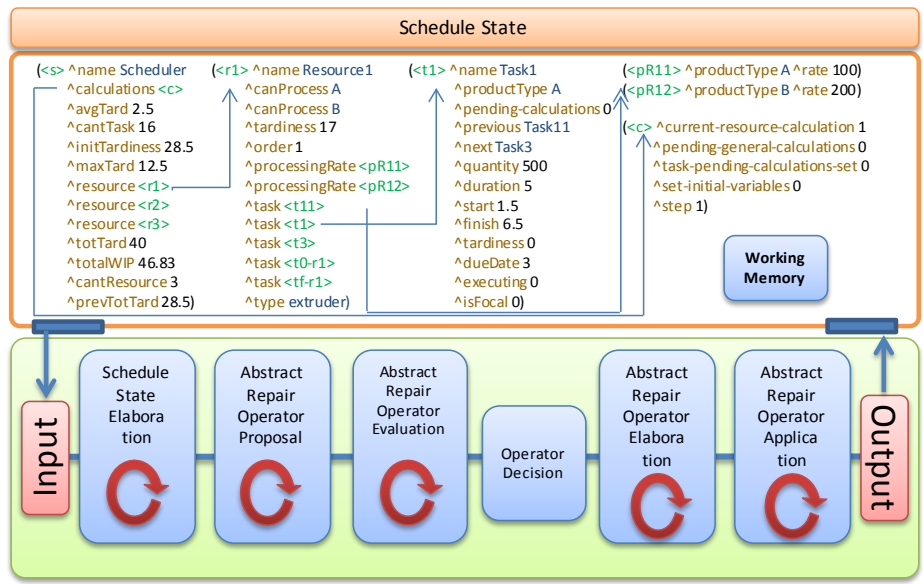

**Fig. 1.** Symbolic representation of schedule state (above) and repair operator elaboration, proposal and application cycle.

The concept of state is a key component to solve instances of rescheduling problems in SOAR, because SOAR programs are implicitly organized in terms of Problem Space Computational Model [14] so as to the conditions for proposing operators will restrict any operator to be considered only when it is relevant, and thus define the space of possible states that might be considered for achieving a repair goal. In this approach, the problem space is comprised of all possible schedules that can be generated when solving rescheduling tasks, and all repair operators that give rise to a schedule transition from one state to another. However, in a rescheduling task the architecture does not explicitly generate all of feasible states exhaustively; instead, SOAR is in a specific schedule state at a given time (represented in working memory), attempting to select an operator to that will move it to a new, hopefully improved state. Such process continues recursively towards a goal state (i.e. schedule with a total tardiness that is minor than the initial tardiness). The lower section of Figure 1 shows the schedule repair process execution, which proceeds through a number of cycles. Each cycle is applied in phases; in the Input phase, new sensory data comes into the working memory and is copied as the actual schedule state. In the elaboration phase production rules fire (and retract) to interpret new data and generate derived facts (state elaboration). In the Proposal phase the architecture propose repair operators for the current schedule state using preferences, and then compare proposed operators (in evaluation phase). All matched production rules fire in parallel (and all retractions occur also in parallel), while matching and firing continue until there are no more additional complete matches or retractions of production rules (quiescence). Therefore, a decision procedure selects a new operator based on numeric preferences provided the reinforcement learning rules. Once a repair operator has been selected, its application phase starts and the operator application rules fire. The actions of these productions give rise to more matches or retract operations; just as during proposal, productions fire and retract in parallel until quiescence. Finally, output commands are sent to the real-time rescheduling component. The cycle continues until the halt action is issued from the Soar program (as the action of a production rule).

### 3.1 Design and Implementation of Repair Operators and Rescheduling Goals

As was explained in the previous section, repair operators are the way by which a schedule goes from one state to another until the rescheduling goal is reached. Hereby, deictic repair operators have been designed to move or swap a focal task with other tasks in the schedule (which may be assigned to other resource), so as to reach a goal state [10]. Each operator takes two arguments: the focal task, and an auxiliary task. Focal task is taken as the reparation anchorage point, and auxiliary task serves to specify the reparation action and evaluate its effectiveness. For example, if the proposed operator is `down-right-jump`, the idea is that the focal task must be moved to an alternative resource, and inserted after the auxiliary task. If so, the conditions of the `down-right-jump` operator proposal rule must assure that the auxiliary task has a programmed start time which is greater than the start time of the focal task before it has been moved. It is important to note that in the alternative resource may exist more than one task that meet the condition; in such case, the operator is proposed in parallel, parameterized with different auxiliary tasks. To exemplify the reasoning above, Figure 2 shows the `down-right-jump` application rule (left hand side at the left, and right hand side at the right). The left hand side of the rule in Figure 2 establishes the conditions that must be met by the schedule state so that the operator can be applied. In turn, the right hand side defines how the schedule state changes as a consequence of the repair operator application. All symbols enclosed in "<>" represent variables, and variables with the same name refer to the same object in both parts of the rule. Therefore, the rule in Figure 2 can be semantically expressed as: "if in the schedule, there exists a proposed operator named, `down-right-jump` which takes as argument a focal task named `<nametFocal>` and auxiliary task `<nametAux>`; also, there exists a resource `<r1>` which has assigned tasks `<tFocal>`, `<tPrevFocal>` and `<tPosFocal>` which are different from each other, there exists a resource `<r2>` with a processing rate of `<rater2>` for the product type `<prodType>` which has assigned the different tasks `<tAux>` and `<tPosAux>`. The task `<tFocal>` is the focal task of the repair operator and its attributes take the values `<quantity>`, `<duration>` and `<prodType>`, the previous programmed task in the resource is `<prevTFocal>` and the next is `<nexttFocal>`. The task `<tAux>` is named `<nametAux>` and it is the auxiliary task in the repair operator and has as a previous programmed task named `<prevtAux>` and a next task `<nexttAux>`. Further, task `<tPrevFocal>` is named `<nametPrevFocal>`, task `<tPosFocal>` is named `<nametPosFocal>` and task `<tPosAux>` is named `<nametPosAux>`, then as a result of the operator application the schedule state change in the following manner: the new previous task of `<tFocal>` is `<nametAux>` and the next task is `<nametPosAux>`, the new next task of `<tAux>` is `<nametFocal>`, the new previous task of `<tPosAux>` is `<nametFocal>`, the new previous task of `<tPosFocal>` is `<nametPrevFocal>`, the new next task of `<tPrevFocal>` is `<nametPosFocal>`. Also, the new duration value of the focal task is `(/ <quantity> <rater2>)`, the focal task is moved to resource `<r2>` and removed from `<r1>` and the value of pending-general-calculations is updated in 1.

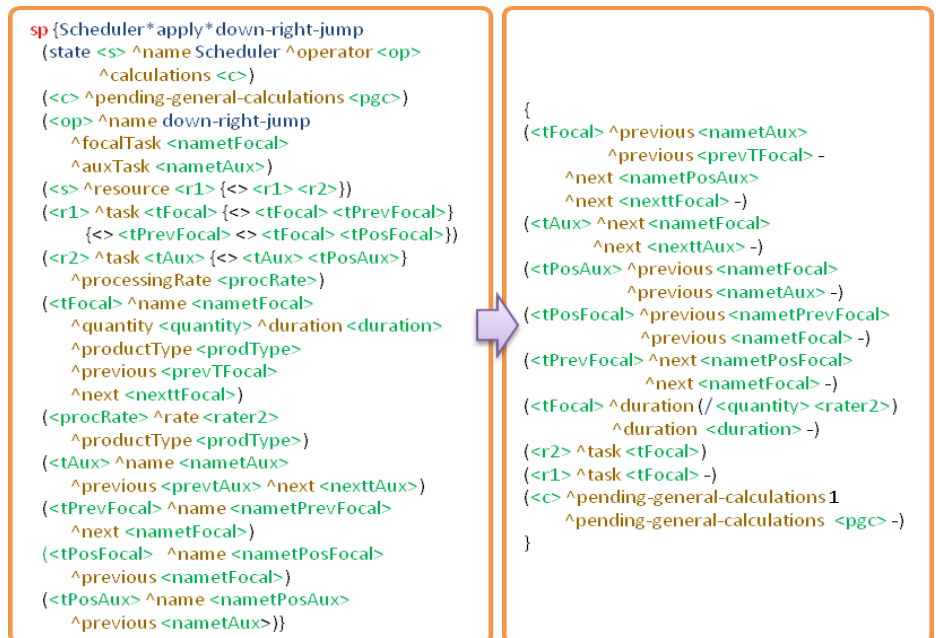

**Fig. 2.** Down-right-jump application rule.

As a result of the rule application the focal task has been moved to an alternative resource, its duration has been recalculated as from the processing rate of the alternative resource, and it has been inserted after the auxiliary task and the task originally programmed to start before of it. The pending-general-calculations value in 1 fires new operators with the aim of recalculating task tardiness, resource tardiness, and other numeric values. An important matter is that once the repair operator and its arguments have been obtained, the rest of the variable values can be totally defined because they are related with each other, allowing an effective repair operator application. Another advantage of this approach relies on the use of variables in the body rule which act as universal quantifier, so the repair operator definition and application can match totally different schedule situations and production types only with the relational restrictions established in the left hand side of the rules.

Finally, after changes in the schedule state have been performed by the repair operator application rule, SOAR reinforcement learning rules are fired, so that the architecture can learn numeric preferences from the results of the particular repair operator application, which is carried on using a reward function and the SARSA(λ) algorithm [14]. The reward function is defined as the amount of tardiness reduced (positive reward) or increased (negative reward). Therefore, the SARSA(λ) algorithm updates the numeric preference of the operator using the well known formula in Eq. (1)

$$Q(s,ro)_{t+1} = Q(s,ro)_t + \alpha[r + \gamma Q(s',ro')_t - Q(s,ro)_t]e(s,ro)_t \qquad (1)$$

where $Q(s,ro)$ is the value of applying the repair operator $ro$ in the schedule state $s$, whereas $\alpha$ and $\gamma$ are algorithm parameters, $r$ is the reward value while $e(s,ro)$ is the eligibility trace value for repair operator $ro$ in state $s$. Because of the problem space can be extremely large, and *Q*-values are stored in production rules which cannot be predefined, a reinforcement learning rule template [11] must be defined in order to generate updateable numeric preference rules that follow SOAR specifications for performing the learning procedure whenever visiting schedule states by means of available repair operators.

## 4 Industrial Case Study

An example problem proposed by Musier and Evans in [15] is considered to illustrate our approach for automated task rescheduling. It consists of a batch plant which is made up of three semi-continuous extruders that process customer orders for four classes of products (A, B, C and D). Each extruder can process one product at a time, and has its own characteristics. For example, the production rate for each type of product may vary from one extruder to another, and each extruder is not necessarily capable to process each type of product. Each task, in turn, has a due date, a required quantity (expressed in Kg.) and a product type.

Three applications have been used to implement and test the case study: *VisualSoar* v4.6.1, *SoarDebugger* 9.3.2 [12] and Visual Studio 2010 Ultimate running under Windows 7. Visual Studio 2010 was used to develop the real time scheduling component which allowed validating the results and read/write on the SOAR output/input link, respectively. *VisualSoar* environment was used to design and implement the definition of schedule state and operator proposal, elaboration and application knowledge.

Furthermore, *SoarDebugger* was used to run the aforementioned rules and train a rescheduling agent as well as to analyze the correctness of the operator pro-posal/application rules. For the rescheduling problem space, there was a maximum of ten repair operators proposed for any state in each repair step and there are two classes of operators: *move* operators, which move the focal task into another position in the same resource or into an alternative one, and *swap* operators, which exchange the focal task with another task on different resources.

After each repair step, if the schedule has been repaired, the architecture is halted; otherwise, the agent propose/apply a new operator until the goal is achieved or an excessive amount of episodes has been made without finding the rescheduling solution. This situation may occur when the schedule to be repaired is very similar to the optimal schedule so further improvements are difficult to obtain Also, for each repair operator application, a reward is given to the agent based on how close the current schedule's tardiness is from the initial tardiness (i. e., how close is the repaired schedule to the goal state).

In this work, the disruptive event which has been considered is the arrival of a new order; for learning, the Sarsa ($\lambda$) algorithm was used with an ε-greedy police, eligibility traces, and parameters: $\gamma = 0.9$, $\varepsilon = 0.1$, $\lambda = 0.1$ and $\alpha = 0.1$. In the training phase of the rescheduling agent 20 simulated episodes were executed. In some episodes, the agent achieved the repair goal, which was stated as "*insert the arriving order without increasing the total tardiness present in the schedule before the order was inserted*". The intention behind such repair goal was on one hand, to insert the new order in an efficient way, and on the other, take advantage of the opportunity to improve schedule efficiency which may be caused by the very occurrence of a disruptive event [2] if it is possible. As the result of the training phase, 2520 reinforcement learning rules was generated dynamically. A representative example of one of such rules can be seen in Example 1 below (Some components of its left hand side have been omitted to facilitate reading and easy semantic interpretation).

```
Example 1. sp {rl*Scheduler*rl*rules*157
    (state <s1> ^totalWIP 46.83 ^taskNumber 16 ^maxTard 15
    ^avgTard 2.5 ^totTard 40 ^initTardiness 28.5 ^name
    Scheduler ^operator <o1> + ^focalTask <t1>)
    (<o1> ^auxTask Task10 ^focalTask Task5 ^name up-right-
    jump) --> (<s1> ^operator <o1> = -0.1498)}
```

The Example 1 rule was automatically instantiated by the SOAR cognitive architecture from an abstract learned rule, and is carried over the next schedule repair operations, so using these learned rules rescheduling decisions can be performed reactively in real time without extra deliberation. The rule in Example 1 reads as follows: if the Schedule state named `<s1>` has a Total Work in Process of 46.83, a Task Number of 16, a Maximum Tardiness of 15, an Average Tardiness of 2.5, a Total Tardiness of 40, an Initial Tardiness of 28.5, a Focal Task `<t1>` and the repair operator applied is up-right-jump, taken as auxiliary Task10 and Focal Task Task5, then the *Q*-value of that repair operator application is -0.1498. Evaluating such values for each operator, the agent can determine which one is the best in each situation, and act accordingly. Once the learning process has been performed, a new schedule was generated to test the learned repair policy.

To generate the initial schedule, 15 orders have been assigned to 3 resources as can be seen in Fig 3. Before the insertion of the focal task (highlighted in white), the schedule total tardiness was 28.5 h. After the insertion of the focal task the schedule total tardiness has increased to 40 h. (see Fig. 4a). After six repair steps using the SOAR knowledge base, the initial schedule was successfully repaired, and the total tardiness was reduced to 18.5 h. (see Fig. 4b). Executing tasks at the time of insertion are highlighted in orange. The sequence of repair operators applied by SOAR was the following: the first operator was `up-right-jump` to Resource 1 (so Focal Task was inserted between Task 3 and Task 2), increasing the total tardiness to 44 h. That was followed by a `down-right-jump` to Resource 2 (behind Task 16), decreasing the total tardiness to 28.5 h. The third operator applied was an `up-right-jump` to Resource 1 (ahead of Task 10), increasing the total tardiness to 46.5 h. The next was a `down-right-jump` to Resource 3 (between Task 14 and Task 4). Next, SOAR applied an `up-left-jump` to Resource 2, (between Task 6 and Task 12). Finally, a `down-left-swap` to R3 was applied with Task 7, leaving the schedule with a total tardiness of 18.5 h.

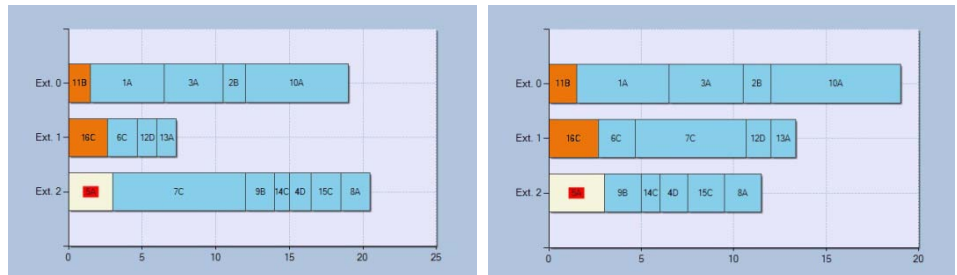

(a). Initial Tardiness: 28.50 h. Tardiness After Insertion: 40.00 h.

(b). Final Tardiness after applying the repair policy: 18.50 h.

**Fig. 3.** Example of applying the optimal sequence of repair operators

As can be seen in the repair sequence, the rescheduling policy tries to obtain a balanced schedule using the focal task as the basis for swapping order positions in sequence of generated schedules, in order to take advantage of extruders with the best processing times. In this case, the rescheduling agent tries to relocate the Focal Task in an alternative resource so as to make a swap with Task 7 in order to move it to the second, sub utilized extruder. It is worth highlighting the small number of steps that are required for the scheduling agent to implement the learned policy in order to handle the insertion of an arriving order.

## 5 Concluding Remarks

A novel approach for simulation-based learning of a rule-based policy dealing with automated repair in real time of schedules using the SOAR cognitive architecture has been presented. The rescheduling policy allows generating a sequence of local repair operators to achieve alternative rescheduling goals which help coping with abnormal and unplanned events such as inserting an arriving order with minimum tardiness based on a symbolic first order representation of schedule states using abstract repair operators. The proposed approach efficiently represents and uses large bodies of symbolic knowledge, because it combines dynamically available knowledge for decision-making in the form of production rules with learning mechanisms. Also, it compiles the rescheduling problem into production rules, so that over time, the schedule repair process is replaced by rule-driven decision making which can be used reactively in real-time in a straightforward way. In that sense, the use of abstract repair operators can match several non-predefined situations representing rescheduling tasks by means of problem spaces and schedule states using a relational symbolic abstraction which is not only efficient to profit from, but also potentially a very natural choice to mimic the human cognitive ability to deal with rescheduling problems, where relations between focal points and objects for defining repair strategies are typically used. Finally, by relying on an appropriate and well designed set of template rules, the approach enables the automatic generation through reinforcement learning and chunking of rescheduling heuristics that can be naturally understood by an end-user.